\documentclass{article}

\usepackage{microtype}
\usepackage{graphicx}
\usepackage{subfigure}
\usepackage{booktabs} 

\usepackage{hyperref}

\usepackage[accepted]{icml2025_arxivChanged}

\usepackage{amsmath}
\usepackage{amssymb}
\usepackage{mathtools}
\usepackage{amsthm}

\usepackage[capitalize,noabbrev]{cleveref}

\usepackage{algorithm}
\usepackage{algorithmic}
\usepackage{multirow}
\usepackage{amssymb}
\usepackage{pifont}
\usepackage{amsmath}  
\usepackage{amssymb}  
\usepackage{bm}    

\usepackage{color,xcolor,xspace}
\usepackage{colortbl}
\usepackage{amsthm}

\theoremstyle{plain}
\newtheorem{theorem}{Theorem}[section]
\newtheorem{proposition}[theorem]{Proposition}

\theoremstyle{definition}

\theoremstyle{remark}
\newtheorem{remark}[theorem]{Remark}
\usepackage[textsize=tiny]{todonotes}

\definecolor{shadecolor}{rgb}{0.92,0.92,0.92}
\usepackage{framed}

\usepackage[textsize=tiny]{todonotes}

\begin{document}

\twocolumn[
\icmltitle{Improved Feature Generating Framework for Transductive Zero-shot Learning}



\icmlsetsymbol{equal}{*}

\begin{icmlauthorlist}
\icmlauthor{Zihan Ye}{XJTLU,UOL,WLU}
\icmlauthor{Xinyuan Ru}{XJTLU,UOL}
\icmlauthor{Shiming Chen}{MBZUAI}
\icmlauthor{Yaochu Jin}{WLU}
\icmlauthor{Kaizhu Huang}{DKU}
\icmlauthor{Xiaobo Jin}{XJTLU}
\end{icmlauthorlist}

\icmlaffiliation{XJTLU}{Xi'an Jiaotong-Liverpool University;}
\icmlaffiliation{UOL}{University of Liverpool;}
\icmlaffiliation{MBZUAI}{Mohamed bin Zayed University of Artificial Intelligence;}
\icmlaffiliation{WLU}{Westlake University;}
\icmlaffiliation{DKU}{Duke Kunshan University}

\icmlcorrespondingauthor{Xiaobo Jin}{Xiaobo.Jin@xjtlu.edu.cn}

\icmlkeywords{Machine Learning, ICML}

\vskip 0.3in
]



\printAffiliationsAndNotice{}  


\begin{abstract}
Feature Generative Adversarial Networks have emerged as powerful generative models in producing high-quality representations of unseen classes within the scope of Zero-shot Learning (ZSL). This paper delves into the pivotal influence of unseen class priors within the framework of transductive ZSL (TZSL) and illuminates the finding that even a marginal prior bias can result in substantial accuracy declines. Our extensive analysis uncovers that this inefficacy fundamentally stems from the utilization of an unconditional unseen discriminator — a core component in existing TZSL. We further establish that the detrimental effects of this component are inevitable unless the generator perfectly fits class-specific distributions.
Building on these insights, we introduce our Improved Feature Generation Framework, termed \textbf{I-VAEGAN}, which incorporates two novel components: Pseudo-conditional Feature Adversarial (PFA) learning and Variational Embedding Regression (VER).
PFA circumvents the need for prior estimation  
by explicitly injecting the predicted semantics as pseudo conditions for unseen classes premised by precise semantic regression.
Meanwhile, VER utilizes reconstructive pre-training to learn class statistics, obtaining better semantic regression.
Our I-VAEGAN achieves state-of-the-art TZSL accuracy across various benchmarks and priors.
Our code would be released upon acceptance.
\end{abstract}


\section{Introduction}
\label{sec:intro}

\begin{table}[htbp]
    \centering
    
\caption{We analyze the relationship between TZSL accuracy and the prior bias (the averaging the absolute error between estimated prior and real prior) on Bi-VAEGAN.
    T1 and PB denote the TZSL top-1 accuracy on unseen classes and class-averaged prior bias, respectively.
    $\dagger$ and $\ddagger$ denote the datasets have a non-uniform or close-uniform distribution of unseen classes.
    We can conclude: for non-uniform unseen class distribution, \textit{small prior bias, big accuracy degradation.} }
    
        \begin{tabular}{c|cc|cc|cc}
        \hline
        \hline
        \multirow{2}{*}{\textbf{Prior}}  & \multicolumn{2}{c}{\textbf{AWA1$^\dagger$}}  & \multicolumn{2}{c}{\textbf{AWA2$^\dagger$}} & \multicolumn{2}{c}{\textbf{CUB$^\ddagger$}} \\
        \cline{2-7}
        & T1 & PB & T1& PB & T1&PB  \\
        \hline
        Ground Truth & 93.9& NA &  95.8& NA & 76.8& NA \\
        Uniform & 66.3& 3.62 &  60.3& 5.67 & 76.8&0.3 \\
        CPE & 91.5 & 1.14 &  85.6 & 2.26 & 74.0& 4.6  \\
        \hline
        \hline
    \end{tabular}

    \label{tab:banner}
\end{table}

Zero-shot learning (ZSL) emerges as a solution aimed at circumventing the extensive annotation and training costs that are inherent to traditional machine learning~\cite{xian2018zero}. The central objective of ZSL is the recognition of ``unseen" classes without any prior exposure, utilizing knowledge exclusively gained from ``seen" classes during training phase. The crux of ZSL is the strategic transfer of semantic knowledge from seen to unseen classes through manually annotated attributes~\cite{xu2022attribute}, or text-embedding methods such as text2vec~\cite{zhu2018a}.

The advent of generative models has markedly influenced the ZSL landscape due to their formidable ability to generate new data points. A variety of studies have delved into the application of Generative Adversarial Networks (GANs)~\cite{ye2021disentangling}, and Variational Auto-Encoders (VAEs)~\cite{chen2021semantics}, or a composite methodology named VAEGANs~\cite{xian2019f} that aims to internalize the visual-semantic distribution of seen classes. The implication of this generative strategy is profound: by synthesizing representations for unseen classes based on their semantic profiles, ZSL transcends its conventional boundaries, morphing into a traditional supervised learning  problem.

Recently, to mitigate the discrepancy between seen and unseen classes, the research community has introduced a novel paradigm known as transductive Zero-shot Learning (TZSL)~\cite{fu2015transductive,wan2019transductive}.  This approach permits the incorporation of unlabeled examples from unseen classes into the training process. The seminal work f-VAEGAN~\cite{xian2019f} leverages these unlabeled samples from unseen classes and innovates by integrating an additional unconditional discriminator, denoted as $D_{u}$, to bridge the divide between the real and synthesized feature distributions for the unseen classes. This unconditional discriminator has subsequently become a cornerstone in contemporary generative TZSL methodologies~\cite{narayan2020latent, marmoreo2021transductive, wang2023bi, liu2024transductive}. Further advancing the field, the study Bi-VAEGAN~\cite{wang2023bi} introduces an efficient Cluster Prior Estimation (CPE) technique that calculates the prior for unseen classes through a clustering approach to analyze examples of these classes.

In this paper, we observe a nuanced phenomenon described as \textit{small prior bias, big accuracy degradation.} We investigate the impact of varying prior types on TZSL accuracy, specifically focusing on CPE and uniform prior, with our findings presented in Table \ref{tab:banner}. Our analysis reveals that within datasets featuring an even distribution of unseen classes, even a small class-averaged prior bias can precipitate a dramatic decline in accuracy. Specifically, at AWA1, the uniform prior  has only $3.62\%$  probability bias per unseen class, but the TZSL accuracy declines by $27.6\%$,  from $93.9\%$ to $66.3\%$. Although CPE reduces the prior bias to some extent, it still induces a large performance gap compared to using Ground Truth (GT) prior.
For example, at AWA2, the CPE prior bias declines from $5.67\%$ (uniform prior) to $2.26\%$, while the TZSL accuracy drops about $10\%$  compared to GT prior.

To elucidate the shortcomings observed, we conducted a series of controlled experiments through which we uncovered a ``prior reaction chain." This investigation reveals that the deficiency is rooted in a key component of f-VAEGAN, denoted as $D_{u}$, which unconditionally maximizes the gap between real and generated feature distributions of unseen classes.
Our theoretical analysis shows that, even if the generator could fully fit the real unconditional feature distribution, the class-conditioned feature distributions still lead to non-negligible discrepancies, since $D_{u}$ only distinguishes real samples from fakes despite their class conditions. In case of a biased estimated prior, the prior bias would be accumulated and harm the judgement of $D_{u}$, thereby imposing incorrect guidance to our generators, and eventually damaging our unseen class generation and classification.

To solve the problem, inspired by the prior reaction chain, we propose our Improved Feature Generating framework (\textbf{I-VAEGAN}), built upon the baseline f-VAEGAN.
Overall, I-VAEGAN covers two key designs: Pseudo-conditional Feature Adversarial (PFA) learning and Variational Embedding Regression (VER).
Differing the traditional unconditional discriminators for unseen classes, PFA judges visual sample features with pseudo class conditions --- by a visual$\to$semantic regressor.
Through explicitly expressing pseudo class conditions, the accumulated prior error is mitigated. While other studies, such as~\cite{yang2024consistency, mohebi2024transductive}, recognize the significance of pseudo conditions, they do not explore  the underlying reasons for their efficacy. In our work, we demonstrate that if the semantic regressor is sufficiently accurate, pseudo conditions can approximate real conditions perfectly, thus circumventing
the need for accurate prior estimation.

Moreover, the semantic regression on unseen classes is often  considered as a domain-adaption problem~\cite{schonfeld2019generalized,chen2021hsva}.
To further improve the semantic regressor, we take a new regression method detached from conventional domain-adaption-based regression: VER.
It pre-trains an additional VAE to model intra-class variance in advance.
During the pre-training, samples of all classes are projected into a more structured latent space and reconstructed from the VAE-embeddings. Then, the projected VAE-embeddings are concatenated with visual features to our regressor to enhance its semantics prediction ability.
Given that VAE pre-training is an unsupervised process, VER can be seamlessly integrated with existing regression methodologies, regardless of whether they are supervised~\cite{narayan2020latent,ye2023rebalanced} or unsupervised~\cite{chen2021free,wang2023bi}.

To summarize, the main contributions of this paper are:
\begin{enumerate}
    \item We prove that even the generative distribution fully fits real distribution under unconditional situation, their class-specific distributions still have non-negligible discrepancies.
    To the best of our knowledge, we are the first that explicitly dissects out a prior reaction chain from the impact of prior bias.
    \item We propose a novel framework I-VAEGAN. Our PFA remedies the impact of prior bias by explicitly injecting pseudo class conditions. Our VER puts forward a new way for domain adaption regression by projecting seen and unseen classes into reconstructive latent space, reducing the regression error.
    \item  Extensive experiments verify that our method outperforms existing methods across diverse benchmarks and priors. 
\end{enumerate}

\section{Related Work}
ZSL~\cite{xian2018zero} relies on auxiliary semantics, which are usually in the form of textual descriptions (word2vec, text2vec)~\cite{reed2016learning, naeem2022i2dformer} or attributes~\cite{xu2022attribute, chen2022transzero}, to fill the gaps between seen and unseen classes.
Existing inductive ZSL approaches could be divided into two dominant types: embedding and generative methods.
Embedding methods~\cite{ye2023rebalanced, xu2022attribute, chen2022transzero} leverage a cross-modality mapping and classify unseen classes into the mapped space.
Generative approaches~\cite{narayan2020latent, zhao2022boosting}, utilize generative models, like Variational Auto-Encoder (VAE)~\cite{chen2021semantics}, Generative Adversarial Network (GAN)~\cite{xian2018feature, ye2019sr},  and Diffusion~\cite{ye2024exploring}, to synthesize pseudo samples from given semantics and train the final ZSL classifier. Despite the usefulness of these methods, since the knowledge of unseen classes is often unavailable, their performance is restricted by the quality of the auxiliary knowledge.

As a compromised setting of inductive ZSL, \textbf{transductive ZSL (TZSL)} allows to use unlabeled examples of unseen classes into training~\cite{fu2015transductive,wan2019transductive}.
The typical TZSL work f-VAEGAN~\cite{xian2019f}  combines VAE and GAN, and designs an unconditional discriminator to model the real distribution of unseen classes.
Another important method, TF-VAEGAN~\cite{narayan2020latent}, leverages a plain semantic regressor to refine generated features for inductive semantics.
Recently, Bi-VAEGAN~\cite{wang2023bi} enhances the semantic regressor by adversarially training the regressor and a semantic discriminator in a transductive setting.
However, these methods ignore the dramatic impact of unseen class prior.
Although Bi-VAEGAN recognizes the importance of unseen class prior and proposes CPE to estimate it, they do not explicitly analyze its impact.
In this paper, we discover a prior reaction chain which inspires us to refine the unconditional unseen discriminator by designing PFA to mitigate the prior bias impact.
Some work also  leverages pseudo conditions~\cite{mohebi2024transductive, yang2024consistency}.
However, they did not study the prior impact sufficiently.
Moreover, our VER improves the plain regressor and adversarial regressor, offering a new way for unseen class semantic regression.

\section{Proposed Method}
\subsection{Preliminary}
\subsubsection{Notation}
To train a TZSL model, we have $n_s$ labeled samples from seen classes $\mathcal{C}^s$ and $n_u$ unlabeled samples from unseen classes $\mathcal{C}^u$.
We use $N_s$ and $N_u$ to mark the number of seen and unseen classes.
These two class sets are disjoint, i.e., $\mathcal{C}^s \cap \mathcal{C}^u = \emptyset$.
We specify $\mathcal{X}^{s} = \{\mathbf{x}^s_i\}^{n_s}_{i=1}$ as the visual features from $\mathcal{C}^s$ and $\mathcal{X}^u = \{\mathbf{x}^u_i\}^{n_u}_{i=1}$ from $\mathcal{C}^u$.
These visual features are extracted by a pre-trained network~\cite{xian2018zero}.
For seen classes, instance-level class labels are denoted by $\mathcal{Y}^s = \{y^s_i\}^{n_s}_{i=1} $ and class-level semantic labels $\mathcal{A}^s = \{\mathbf{a}^s_i\}^{n_s}_{i=1} $.
For unseen classes, we cannot access instance-level class labels, but we can use unseen semantic labels $\mathcal{A}^s = \{\mathbf{a}^u_i\}^{N_u}_{i=1} $.
In short, the training set is $D^{tr} = \{\langle \mathcal{X}^{s}, \mathcal{Y}^s\rangle, \mathcal{X}^{u}, \{\mathcal{A}^s, \mathcal{A}^u\} \}$, where $\langle \cdot, \cdot \rangle$ highlights paired data.

\subsubsection{f-VAEGAN}
The milestone work f-VAEGAN combines VAE and GAN to improve the training stability and generation performance.
It consists of a feature encoder $E$, a feature generator (also a decoder) $G$, and two discriminators: a conditional $D_{s}$ for seen classes and an unconditional $D_{u}$ for unseen classes.
The encoder $E$ projects a pair of feature and semantic label from seen classes to a latent representation, i.e., $ [\mathbf{\mu}^s, \sigma^s] = E(\mathbf{x}^{s}, \mathbf{a}^{s})$.
$\mathbf{\mu}^s$ and $\sigma^s$ denote the mean and the log of variance to re-parameterize the embedding distribution.
Then,  a latent representation $\tilde{\mathbf{z}}^{s}\sim\mathcal{N}(\mathbf{\mu}^s,\mathbf{\sigma}^s)$ is sampled, and $G$ decodes $\tilde{\mathbf{z}}^{s}$ to reconstruct visual feature, i.e., $\tilde{\mathbf{x}}^{s} = G(\tilde{\mathbf{z}}^{s}, \mathbf{a}^{s})$ by optimizing:
\begin{align}
\label{eq:fvaegan-vae}
\mathcal{L}^{s}_{VAE} & = KL\left( \mathcal{N}(\mathbf{\mu}^s,\mathbf{\sigma}^s) \| \mathcal{N}(\mathbf{0},\mathbf{1}) \right) \nonumber \\
&+ \mathbb{E} [\| G(\tilde{\mathbf{z}}^s, \mathbf{a}^{s}) -  \mathbf{x}^{s}\|^2_2 ],
\end{align}
where the first term is the Kullback-Leibler divergence and the second is the reconstruction term by mean-squared-error (MSE).
On the other hand, conditional $D_{s}$ distinguishes seen features $\mathbf{x}^{s}$ from generated features $\tilde{\mathbf{x}}^{s}$ with respect to the semantic label $\mathbf{a}^{s}$.
The explicit optimization loss is
\begin{align}
\label{eq:fvaegan-gan-s}
    \mathcal{L}^{s}_{GAN} &= \mathbb{E}[D_s(\mathbf{x}^s, \mathbf{a}^{s})] - \mathbb{E}[D_s(\tilde{\mathbf{x}}^s, \mathbf{a}^{s})] \nonumber \\
    & - \lambda_{gp} \mathbb{E}[(\|\nabla_{\hat{\mathbf{x}}^s} D_s(\hat{\mathbf{x}}^s, \mathbf{a}^{s})\|_2 - 1)^2],
\end{align}
where $\hat{\mathbf{x}}^s = \alpha \mathbf{x}^s + (1 - \alpha) \tilde{\mathbf{x}}^s$ with $\alpha \sim U(0, 1)$ and $\lambda_{gp}$ is a coefficient for the gradient penalty term~\cite{gulrajani2017improved}.
Besides, to leverage unlabeled unseen samples $\mathbf{x}^u$, the unconditional $D_{u}$ also participates in the adversarial training:
\begin{align}
\label{eq:fvaegan-gan-u}
    \mathcal{L}^{u1}_{GAN} &= \mathbb{E}[D_u(\mathbf{x}^u)] - \mathbb{E}[D_u(\tilde{\mathbf{x}}^u)] \nonumber \\
    & - \lambda_{gp} \mathbb{E}[(\|\nabla_{\hat{\mathbf{x}}^u} D_u(\hat{\mathbf{x}}^u)\|_2 - 1)^2],
\end{align}
while $\tilde{\mathbf{x}}^{u} = G(\mathbf{z}, \mathbf{a}^{u})$ with $\mathbf{z} \sim \mathcal{N}(\mathbf{0},\mathbf{1})$ .

\subsection{Prior Reaction Chain}

\begin{table}[thp]
    \centering
    \footnotesize
    \renewcommand{\multirowsetup}{\centering}

\caption{Separately training $D_{u}$ and $G$ with different unseen class prior.
    T1 denotes the TZSL top-1 accuracy on unseen classes.
    $\dagger$ and $\ddagger$ denote the datasets have a non-uniform or close-uniform distribution of unseen classes. We can find $D_{u}$ is far more sensitive than $G$ for varied priors. }
    
    \begin{tabular}{c|c|c|c|c}
    \hline
    \hline
    \multirow{2}{*}{\textbf{$G$ Prior}}  & \multirow{2}{*}{\textbf{$D_u$ Prior}} & \multicolumn{1}{c}{\textbf{AWA1$^\dagger$}}  & \multicolumn{1}{c}{\textbf{AWA2$^\dagger$}} & \multicolumn{1}{c}{\textbf{CUB$^\ddagger$}} \\
    \cline{3-5}
    & & T1 & T1 & T1  \\
    \hline
    Ground Truth & Ground Truth & 93.9  & 95.8  & 76.8 \\
    \hline
    Ground Truth & Uniform & 70.4 & 75.6 & 76.2 \\
    Uniform & Ground Truth & 93.1 & 86.2 & 75.9 \\
    \hline
    Ground Truth & CPE & 91.6 & 87.7 & 72.8 \\
    CPE &Ground Truth & 92.3 & 86.9 & 76.1 \\
    \hline
    \hline
\end{tabular}
    
    \label{tab:PRC}
\end{table}

\begin{figure*}[htbp]
    \centering
    \includegraphics[width=\linewidth]{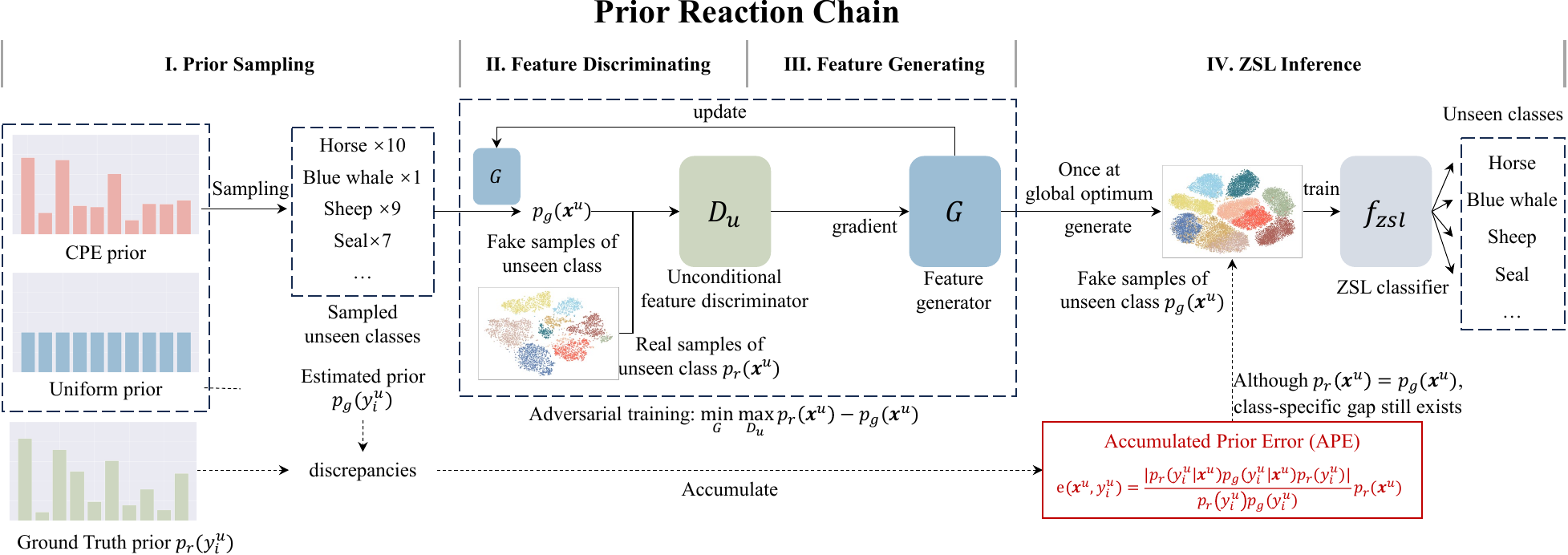}
    \caption{Our discovered prior reaction chain. We find that unseen prior probabilities firstly impact unconditional unseen discriminator $D_{u}$. Then, $D_{u}$ gives insufficient gradient guidance to generator $G$. Next, although unconditional generation distribution $p_g(\mathbf{x}^u)$ could fully fit to real unconditional $p_r(\mathbf{x}^u)$, class-specific distributions still have an inevitable gap. Finally, the ZSL classifier mis-classifies test samples. This chain enlightens us to refine the unconditional $D_{u}$. }
    \label{fig:chain}
\end{figure*}

\subsubsection{Prior Bias Impact}
Since we do not know the prior distribution of unseen classes, we need use some methods to sample unseen classes, i.e. sample $\mathbf{a}^{u}$ in Eq.~\ref{eq:fvaegan-gan-u}.
Typically, we can assume the unseen class distribution is uniform, or estimate the prior by BBSE~\cite{lipton2018detecting} or CPE~\cite{wang2023bi} and let generative models synthesize fake samples according to the assumed prior $p_g(y^{u})$.
However, if the prior estimation is inaccurate, even a small bias may lead to a dramatic impact.
Thus, a question is naturally raised: \textit{how does the prior bias impact TZSL approaches?}

From f-VAEGAN, we have known that the assumed unseen prior $p_g(y^{u})$ determines the sampled unseen classes; $G$ and $D_{u}$ are trained around these sampled unseen classes. Finally, fixing $G$, we train the final $f_{zsl}$.
Thus, we get $p_g(y^{u}) \to G / D_{u} $ and $G \to  f_{zsl} $.
However, since $G$ and $D_{u}$ are alternatively trained, which matters more still remains unclear.

To unravel this, we take a controlled experiment: For transductive training, we use uniform prior to train $D_{u}$ while using GT prior to train $G$. Then, we switch their prior types and re-evaluate the performance.
The results are in Tab.~\ref{tab:PRC}.
We can find that $D_{u}$ is far more sensitive than $G$ for changed priors. 
Applying uniform prior to $D_{u}$ in the first two datasets degrades the performance significantly, while using CPE obtains compensation to some degree.
In contrast, applying the uniform prior to $G$ performs far better than the counterpart to $D_{u}$. Thus, the compensation from CPE is reduced.

\subsubsection{Accumulated Prior Error}
With further analysis, we propose the APE proposition.
Let us only consider $D_{u}$ and $G$.
They learn the objective:
\begin{equation}
    \min_{G} \max_{D_{u}} \quad \mathcal{L}^{u1}_{GAN}.
\end{equation}
When the minimax game achieves its global optimum, the real probability of unseen class samples $p_r(\mathbf{x}^{u})$ is equal to the generated probability $p_g(\mathbf{x}^{u})$~\cite{goodfellow2014generative}.
However, we have the proposition:
\begin{shaded}
    \begin{proposition}[APE]
    When G gets the expected global optimum of the minimax game, i.e., $ p_r(\mathbf{x}^{u}) = p_g(\mathbf{x}^{u})$, for any unseen class $y^u_{i}$, 
        \begin{align}
        \label{eq:chain2}
            {\color{red} e(\mathbf{x}^{u}, y^u_i) } = p_r(\mathbf{x}^{u}|y^u_{i}) - p_g(\mathbf{x}^{u}|y^u_{i}),
        \end{align}
    where $e(\mathbf{x}^{u}, y^u_i)$ is the accumulated prior error:
        \begin{align}
        e(\mathbf{x}^{u}, y^u_i) = \frac{ | p_r(y^u_i|\mathbf{x}^u)p_g(y^u_i) - p_g(y^u_i|\mathbf{x}^u)p_r(y^u_i) | }{p_r(y^u_i)p_g(y^u_i)}  p_r(\mathbf{x}^u). \nonumber
        \end{align}
    \end{proposition}
\end{shaded}
  
\begin{proof}
    The proof is provided in \textbf{Appendix~\ref{appendix:APE}}.
\end{proof}

\begin{remark}
$p_r(y^u_{i}|\mathbf{x}^{u})$ is the real unseen-class-posterior probability, while $p_g(y^u_{i}|\mathbf{x}^{u})$ is the generated unseen-class-posterior probability.
    The $e(\mathbf{x}^{u}, y^u_i)$ becomes $0$ if $| p_r(y^u_i|\mathbf{x}^u)p_g(y^u_i) - p_g(y^u_i|\mathbf{x}^u)p_r(y^u_i) | = 0$, i.e.,
    \begin{align}
    \frac{p_r(y^u_i|\mathbf{x}^u)}{p_g(y^u_i|\mathbf{x}^u) } = \frac{p_r(y^u_i)}{p_g(y^u_i)}.
    \end{align}
    In other words, only when the ratio of the classification probability $p_r(y^u_i|\mathbf{x}^u)$ (trained by real data) to $p_g(y^u_i|\mathbf{x}^u)$ (trained by fake generated data) are \textbf{perfectly} in proportion to the ratio of the real class probability $p_r(y^u_i)$ to the estimated $p_g(y^u_i)$.
    However, existing generation methods still far away from this point as the quantitative comparison of APE we provided in experiments.
    Therefore, the issue of unconditional $D_{u}$ cannot be avoided, showing that $D_{u}$ might be the key factor.
    To this end, a prior reaction chain is emerged, as shown in Fig.~\ref{fig:chain}:
    \begin{align}
    \label{eq:chain}
        p_g(y^{u}) \to D_{u} \to G \to f_{zsl}.
    \end{align}
\end{remark}

\begin{figure}[htbp]
    \centering
    \includegraphics[width=\linewidth]{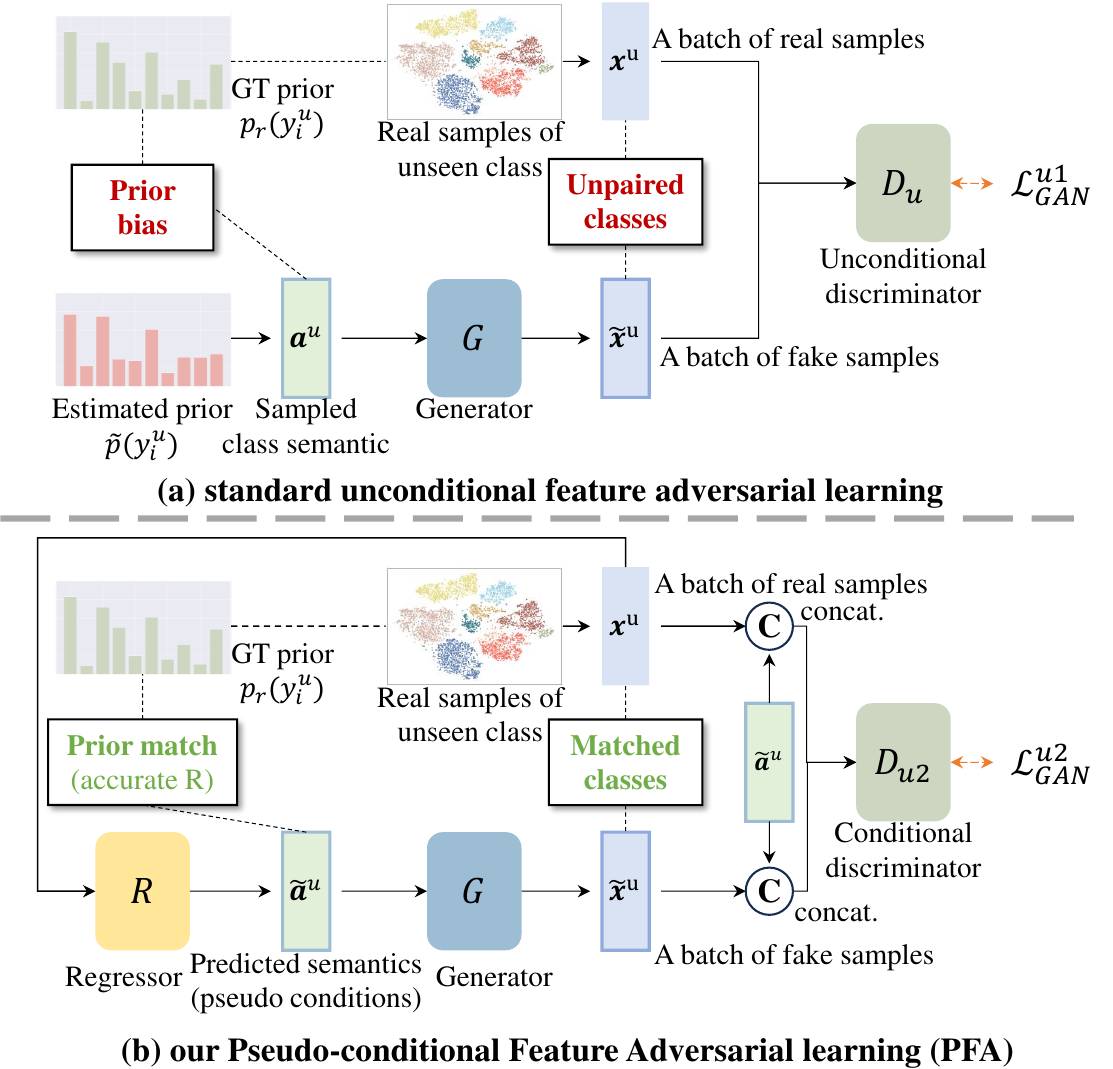}
    \caption{Illustration of our proposed PFA. (a) Standard unconditional feature adversarial learning can be negatively affected by prior bias and unpaired classes of real and fake samples, while (b) our PFA mitigates these two problems: prior and classes can be matched at the same time, only if $R$ is accurate enough. }
    \label{fig:pfa}
\end{figure}

\subsection{Pseudo-conditional Feature Adversarial Learning}
As discussed,  we find that the main cause of the prior chain lies at the fundamental component $D_{u}$, since it does not consider class-specific distributions for feature discriminating.
A natural solution is how to inject class conditions.
In retrospect of $D_{s}$ that judges the pairs of semantics and seen features (Eq.~\ref{eq:fvaegan-gan-s}), can we also do this for unseen features?

To this end, we propose the Pseudo-conditional Feature Adversarial (PFA) learning: we additionally train a semantic regressor $R$ to obtain the pseudo semantic label $\tilde{\mathbf{a}} = R(\mathbf{x}^u)$ and treat it as pseudo class condition for the unseen feature discriminator.
We denote the updated conditional unseen feature discriminator as $D_{u2}$.
Its training loss is: 
\begin{align}
\label{eq:pfa}
    \mathcal{L}^{u2}_{GAN} &= \mathbb{E}[D_{u2}(\mathbf{x}^u, \tilde{\mathbf{a}}^u)] - \mathbb{E}[D_{u2}(\tilde{\mathbf{x}}^u, \tilde{\mathbf{a}}^u)] \nonumber \\
    & - \lambda_{gp} \mathbb{E}[(\|\nabla_{\hat{\mathbf{x}}^u} D_{u2}(\hat{\mathbf{x}}^u, \tilde{\mathbf{a}}^u)\|_2 - 1)^2].
\end{align}

It is worth mentioning that Eq.~\ref{eq:pfa} is not only a conditional version of Eq.~\ref{eq:fvaegan-gan-u}.
As shown in Fig.~\ref{fig:pfa} (a), Eq.~\ref{eq:fvaegan-gan-u} has two drawbacks.
1). It need use the estimated prior to sample semantics and synthesize fake samples; this is impacted by the prior bias.
2). Since the classes of $\mathbf{x}^{u}$ are unavailable, the classes of real features and synthesized features in Eq.~\ref{eq:fvaegan-gan-u} are not paired, which hinders the model optimization.

In contrast, as shown in Fig.~\ref{fig:pfa} (b), our PFA can alleviate these two problems.
First, since the generation condition  $\tilde{a}^{u}$ comes from real samples,  $\tilde{a}^{u}$ can match the ground truth prior, if only the regressor is accurate enough.
Second, still due to $\tilde{a}^{u}$, the classes of generated samples are paired with real samples.
In other words, we simply need  think about improving our $R$ instead of the prior bias problem. The reason is that once our $R$ is accurate enough, the class distributions of pseudo conditions can closely approximate the ground truth class distributions infinitely.

\subsection{Variational Embedding Regressor}
\begin{figure}[htbp]
    \centering
    \includegraphics[width=\linewidth]{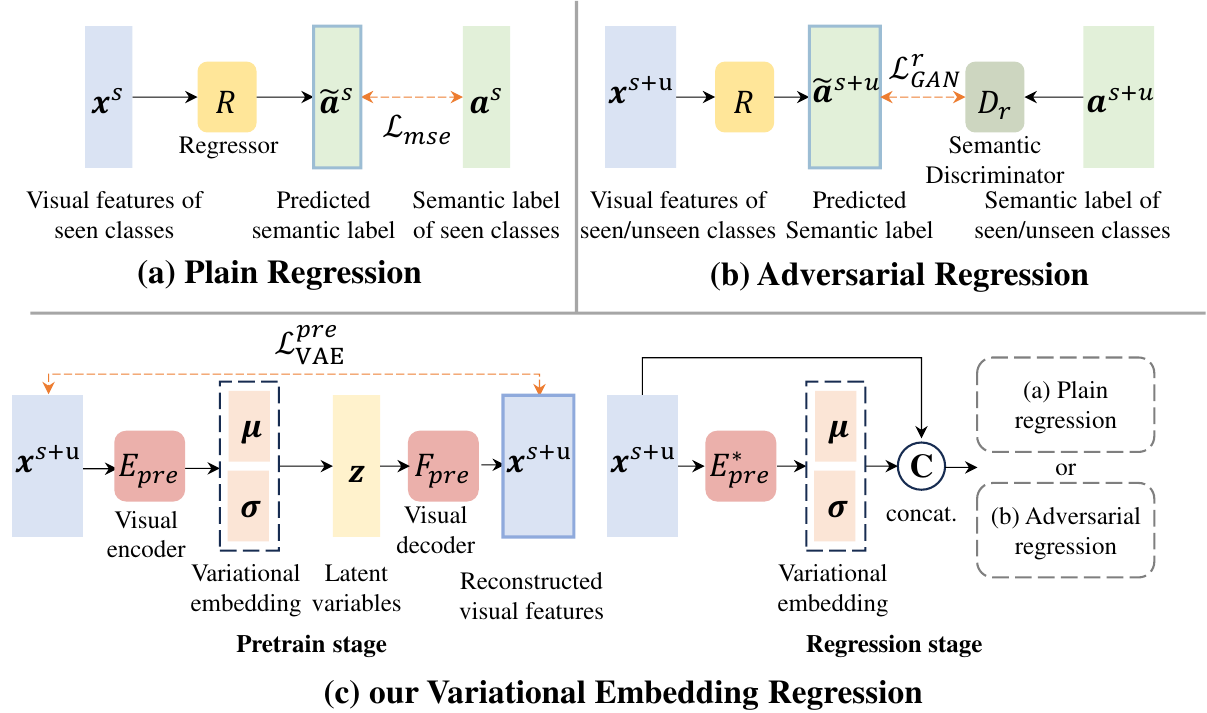}
\caption{Illustration of our VER. (a) Plain regression~\cite{narayan2020latent,ye2023rebalanced} need paired visual features and semantic labels. Thus, it only works on seen classes. (b) Adversarial Regression~\cite{wang2023bi} can work on both seen and unseen classes. (c) Our VER unsupervisedly pre-trains a VAE to model intra-class variations, then uses its embeddings to enhance (a) and (b) seamlessly.}
    \label{fig:ver}
\end{figure}

In this section, we discuss two typical semantic regressors in previous works and our proposed VER. As shown in Fig.~\ref{fig:ver}, the first type is simply minimizing the Mean Squared Error (MSE) loss~\cite{narayan2020latent} or weighted MSE~\cite{ye2023rebalanced} between predicted and real semantic labels:
\begin{equation}
\mathcal{L}_{mse} =\|  \mathbf{a}^s - R(\mathbf{x}^s) \|^2 .
\end{equation}
As MSE is a supervised loss, it can just work on seen classes.
To enhance regression on unseen classes, Bi-VAEGAN proposes the adversarial regressor.
Referring the idea from adversarial training~\cite{wang2023bi}, it utilizes an additional semantic discriminator $D_r$ to distinguish real/fake semantic labels.
As adversarial training is unsupervised, it could work on both seen and unseen classes.
\begin{align}
    \label{eq:gan_r}
    \mathcal{L}^{r}_{GAN} & = \mathbb{E}[D_r(\mathbf{a}^u)] - \mathbb{E}[D_r(\tilde{\mathbf{a}}^u)] \nonumber \\
    & - \lambda_{gp} \mathbb{E}[(\|\nabla_{\hat{\mathbf{a}}^u} D_r(\hat{\mathbf{a}}^u)\|_2 - 1)^2].
\end{align}
However, like other adversarial learning methods, the training of adversarial regression is unstable.

Having acknowledged these progresses, to further improve semantic regression, we propose our VER.
First, previous ZSL works suggests that the original feature space $\mathbf{X}$ might be not distinguished enough.
Thus, they project $\mathbf{X}$ to another space, such as contrastive space~\cite{han2021contrastive} and semantic-disentangling~\cite{chen2021semantics} space.
However, in TZSL, as we cannot get the label of test samples $\mathbf{x}^{u}$, these embedding methods are infeasible.
Inspired by the recent transfer learning pattern:  pre-training, then fine-tune~\cite{radford2021learning, wang2023improving}, we consider introducing an additional VAE to learn the low-dimensional representations of $\mathbf{x}^{u}$ and $\mathbf{x}^{s}$ via unsupervised reconstruction, hoping to provide better embedding.
Specifically, denoting a visual encoder $E_{pre}$ and a decoder $F_{pre}$, 
we pretrain them via a plain VAE loss:
\begin{align}
\label{eq:ver}
\mathcal{L}^{pre}_{VAE} & = KL\left( \mathcal{N}(\mathbf{\mu}_{pre}, \mathbf{\sigma}_{pre}) \| \mathcal{N}(\mathbf{0},\mathbf{1}) \right) \nonumber \\
&- \mathbb{E}_{ \tilde{\mathbf{z}} \sim \mathcal{N}(\mathbf{\mu}_{pre}, \mathbf{\sigma}_{pre}) } [\| F_{pre}(\tilde{\mathbf{z}}) -  \mathbf{x}^{s/u}\|^2_2 ].
\end{align}
After pre-training, we fix $E_{pre}$, and take its variational embedding to construct a low-dimensional latent representation space for all classes. Then,  in the next regression, the variational embedding $\mathbf{\mu}_{pre}, \mathbf{\sigma}_{pre}$ is concatenated with the original visual features.
Then, VER uses the concatenation $[ \mathbf{x}, \mathbf{\mu}_{pre}, \mathbf{\sigma}_{pre} ]$ to replace the original features, seamlessly integrating with the other two types of regressions.

\subsection{Optimization and Inference}
Our I-VAEGAN extends the two-stage training of Bi-VAEGAN to three stages.
Our first stage is our VER pretraining.
And the second stage is regressor training with our VER.
The last stage is generator training with our PFA.
Formally, the overall objective function of I-VAEGAN is formulated as:
\begin{align}
    \label{eq:overall}
    & \text{For stage-1: } \min_{E_{pre}, F_{pre}} \mathcal{L}^{pre}_{VAE}. \nonumber \\
    & \text{For stage-2: } \min_{R} \max_{D_{r}} \mathcal{L}_{mse}+\lambda_{r} \mathcal{L}^{r}_{GAN}. \\
    & \text{For stage-3: } \min_{E,G} \max_{D_s,D_u,D_{u2}} (L_{VAE}^s +L_{GAN}^s  + L_{GAN}^u), \nonumber
\end{align}
$L_{GAN}^u = \lambda_{u1} L_{GAN}^{u1}  +\lambda_{u2} L_{GAN}^{u2}$ and $ \lambda_{u1} $, $ \lambda_{u2} $, $ \lambda_{r} $ is the hyper-parameters indicating the effect of $ L_{GAN}^{u1}$. $ L_{GAN}^{u2}$ and $ L_{GAN}^{r}$  are designed towards the generator.
we also provide the full training algorithm in \textbf{Appendix~\ref{appendix:alg}}.

Finally, after the generator training converges, a classifier $f_{zsl}$ is trained by the synthetic visual features of the unseen class.
We also follow~\cite{narayan2020latent, wang2023bi} to adopt a multi-modal classifier
\begin{align}
\label{eq:inference}
    f_{zsl}: \mathcal{X}^u \times \mathcal{S}^u = \mathcal{X}^u \times \mathcal{A}^u \times \mathcal{H}^u \rightarrow \mathcal{Y}^u.
\end{align}
$\mathcal{H}^u$ is the hidden representations the first fully-connected layer of $R$.
Integrating knowledge of visual space $\mathcal{X}^u$ and semantic space $\mathcal{S}^u$ presents stronger discriminability.

\section{Experiments}


\begin{table}[htbp]
    \centering
    \renewcommand{\multirowsetup}{\centering}
    \caption{Performance comparison in TZSL using different unseen class priors.}
        \begin{tabular}{lp{0.8cm}<{\centering}p{0.8cm}<{\centering}p{0.8cm}<{\centering}p{0.8cm}<{\centering}}
            \hline
            \hline
            Method &  AWA1$^\dagger$& AWA2$^\dagger$ & CUB$^\ddagger$& SUN$^\ddagger$ \\
            \hline
            \multicolumn{5}{l}{\textit{Generative with uniform prior}} \\
            f-VAEGAN & 62.1 & 56.5 & 72.1 & 69.8 \\
            TF-VAEGAN & 63.0 & 58.6 & 74.5 & 71.1 \\
            Bi-VAEGAN & 66.3 & 60.3 & \underline{76.8} & \underline{74.2} \\
            \textbf{I-VAEGAN (Ours)} & 67.0  & 62.7  &  \textbf{77.5}  &  \textbf{75.8} \\
            \hline
            \multicolumn{5}{l}{\textit{Generative with CPE prior}} \\
            Bi-VAEGAN & \underline{91.5} & \underline{85.6} & 74.0 & 71.3 \\
            \textbf{I-VAEGAN (Ours)} & \textbf{92.6}  &   \textbf{86.5} & 74.5  & 73.1  \\
            \hline
            \hline
        \end{tabular}
    \label{tab:unknown}
\end{table}

\begin{table}[thp]
  \centering
   \renewcommand{\multirowsetup}{\centering}

\caption{
  Overall comparison with state-of-the-arts (SOTAs) in TZSL.
  Symbols “I”, “T”, “E”, and “G” indicate inductive, transductive, embedding, and generative ZSL methods, respectively.
  The symbol “$*$” indicates that the methods adopt GT unseen class prior.
  The best and second best results of transductive approaches are marked with \textbf{bold} and \underline{underline}.
  }
   
   \begin{tabular}{p{0.1cm}|p{0.1cm}|p{2.2cm}|p{0.8cm}<{\centering}p{0.8cm}<{\centering}p{0.8cm}<{\centering}p{0.8cm}<{\centering}}
   \hline
      \hline
      \cline{4-7}
      \multicolumn{3}{c|}{Method}&AWA1$^\dagger$&AWA2$^\dagger$&CUB$^\ddagger$&SUN$^\ddagger$\\
      \hline
      \multirow{5}{*}{I}&\multirow{3}{*}{E}&APN&-& 68.4& 72.0 &61.6 \\
 
      & &TransZero++&-& 72.6&78.3& 67.6 \\ 
      
      & &ReZSL&-& 70.9 & 80.9 & 63.2 \\
      
      \cline{2-7}
      &\multirow{2}{*}{G}&f-CLSWGAN&59.9&62.5&58.1 &54.9\\
      
      & &LisGAN&70.6& -& 61.7 &58.8 \\
      
      
      \hline
      
      \multirow{8}{*}{T} & \multirow{2}{*}{E}&PREN&-&78.6&66.4&62.8\\
      
      & &GXE&89.8&83.2&61.3&63.5\\
      
      \cline{2-7}
       & \multirow{5}{*}{G} &f-VAEGAN$^*$&-&89.8&71.1&70.1\\
       
      & &TF-VAEGAN$^*$&-&92.6&74.7&70.9\\
      
      & &SDGN$^*$&92.3&93.4&74.9&68.4\\
      
      & &Bi-VAEGAN$^*$ &\underline{93.9}&\underline{95.8}&\underline{76.8}&\underline{74.2}\\
      
      & & \textbf{I-VAEGAN$^*$} & \textbf{94.4}&\textbf{95.9}&\textbf{77.2}&\textbf{76.0}\\
    
     \hline
     \hline
          \end{tabular}\\
  
    \label{tab:overall_zsl}
\end{table}

\begin{table*}[thp]
  \centering
   \renewcommand{\multirowsetup}{\centering}

\caption{
  Overall comparison with SOTAs for TGZSL.
  The symbols “I”, “T”, “D”, "E" and "G" indicate inductive, transductive, dataset-level, embedding and generative ZSL methods, respectively.
  The symbol “$*$“ indicates the methods adopts GT unseen class prior.
  In GZSL, U, S and H represent the top-1 accuracy (\%) of unseen classes, seen classes, and their harmonic mean, respectively.
  The best and second best results of transductive approaches are marked with \textbf{bold} and \underline{underline}.
  }
   
   \begin{tabular}{l|l|l|ccc|ccc|ccc|ccc}
   \hline
      \hline
      \multicolumn{3}{c|}{\multirow{2}{*}{Method}}&\multicolumn{3}{c}{AWA1$^\dagger$}&\multicolumn{3}{|c}{AWA2$^\dagger$}&\multicolumn{3}{|c|}{CUB$^\ddagger$}&\multicolumn{3}{c}{SUN$^\ddagger$}\\
      \cline{4-15}
      \multicolumn{3}{l|}{}&U&S&H&U&S&H&U&S&H&U&S&H\\
      \hline
      \multirow{5}{*}{I}&\multirow{3}{*}{E}&APN& -&-&-&56.5&78.0&65.5 
 &65.3&69.3&67.2 &41.9&34.0 &37.6 \\ 
 
      & &TransZero++ & -& -& -& 64.6 & 82.7 & 72.5  & 67.5 & 73.6 & 70.4  & 48.6 & 37.8 & 42.5 \\ 
      
      & &ReZSL & - & - &- & 63.8 & 85.6 & 73.1 & 72.8 & 74.8 & 73.8 & 47.4 & 34.8 & 40.1 \\
      
      \cline{2-15}
      &\multirow{3}{*}{G}&f-CLSWGAN& 76.1&16.8&27.5&14.0&81.8&23.9&21.8&33.1&26.3&23.7&63.8&34.4\\
      
      & &LisGAN& 76.3&52.6&62.3&-&-&-&42.9&37.8&40.2&46.5&57.9&51.6 \\
      
      & & f-CLSWGAN+APN& -&-&- & 63.2 & 81.0  & 71.0 & 65.5 & 75.6 & 70.2 & 41.4 & 89.9 & 56.7 \\
      
      \hline
      
      \multirow{9}{*}{T} & \multirow{3}{*}{E}&PREN&-&-&-&32.4&88.6&47.4&35.2&55.8&43.1&35.4&27.2&30.8\\
      
      & &GXE&87.7&\textbf{89.0}&88.4&80.2&90.0&84.8&57.0&68.7&62.3&45.4&\textbf{58.1}&51.0\\
      
      & &PLR &-&-&-&77.2&87.7&82.1&66.4&63.7&65.0&63.2&43.1&51.3\\
      
      \cline{2-15}
       & \multirow{5}{*}{G} &f-VAEGAN$^*$&-&-&-&84.8&88.6&86.7&61.4&65.1&63.2&60.6&41.9&49.6\\
       
      & &TF-VAEGAN$^*$&-&-&-&87.3& 89.6&88.4&69.9&\underline{72.1}&71.0&62.4&\underline{47.1}&53.7\\
      
      & &SDGN$^*$&87.3&88.1&87.7&88.8&89.3&89.1&69.9&70.2&70.1&62.0&46.0&52.8\\
      
      & &Bi-VAEGAN$^*$ &\textbf{89.8}&88.3&\textbf{89.1}&\underline{90.0}&\underline{91.0}&\underline{90.4}&\underline{71.2}&71.7&\underline{71.5}&\underline{66.8}&45.4&\underline{54.1}\\
      
      & & \textbf{I-VAEGAN$^*$ (ours)}&\underline{89.5}&\underline{88.6}&\textbf{89.1}&\textbf{91.2}&\textbf{91.3}&\textbf{91.2}&\textbf{71.8}&\textbf{72.3}&\textbf{72.1}&\textbf{68.8}&46.4&\textbf{55.4}\\
      \hline
      
      \multirow{3}{*}{D}&\multirow{1}{*}{E}&CLIP& -&-&- &88.3 & 93.1& 90.6 & 54.8 & 55.2 &55.0 &-&-& - \\
      \cline{2-15}
       & \multirow{2}{*}{G}  &f-VAEGAN+SHIP& -&-&- &61.2&95.9&74.7 &22.5&82.2& 35.3 & -& -& -\\
       & &TF-VAEGAN+SHIP& -&-&- &43.7&96.3&60.1 & 21.1& 84.4 & 34.0 & -& -& -
       \\
     \hline
     \hline
          \end{tabular}\\
  
    \label{tab:overall_gzsl}
\end{table*}

\subsection{Experiment Settings}
\subsubsection{Datasets}
To demonstrate the powerful TZSL ability of our I-VAEGAN, we conduct experiments using four popular benchmark datasets, including two \textbf{non-uniform ($\dagger$)} animal datasets: AWA1~\cite{lampert2013attribute} and AWA2~\cite{xian2018zero} and two \textbf{close-uniform ($\ddagger$)} datasets: CUB~\cite{wah2011caltech} for birds and SUN~\cite{patterson2012sun} for scene.
More details are provided in Appendix~\ref{appendix:addDataset}.



\subsubsection{Evaluation Metrics}
Under the TZSL setup, we calculate the top-1 classification accuracy average per unseen classes (T1).
We also perform evaluations on a more challenging task, Generalized TZSL (TGZSL), i.e., additional samples of seen classes are included in testing.
For TGZSL, we calculate three kinds of top-1 accuracies, namely the accuracy for unseen classes (U), the accuracy for seen classes (S), and their harmonic mean $H = \frac{2\times U\times S}{U+S}$.

\subsubsection{Implementation Details}
The implementation details are provided in \textbf{Appendix~\ref{appendix:imp}}.

\subsection{Comparison to SOTAs}
\label{sec:sota}
\subsubsection{Baseline}
The compared SOTAs include embedding methods: APN~\cite{xu2020attribute}, TransZero++~\cite{chen2022transzero++}, ReZSL~\cite{ye2023rebalanced},
PREN~\cite{ye2019progressive},
GXE~\cite{li2019rethinking},
PLR~\cite{mohebi2024transductive}.
Generative methods: f-CLSWGAN~\cite{xian2018feature},
LisGAN~\cite{li2019leveraging}, f-CLSWGAN+APN~\cite{xu2022attribute},
f-VAEGAN~\cite{xian2019f},
TF-VAEGAN~\cite{narayan2020latent},
SDGN~\cite{wu2020self},
Bi-VAEGAN~\cite{wang2023bi}.
Besides, the large-scale vision-language models, i.e. CLIP~\cite{radford2021learning} and SHIP~\cite{wang2023improving}, are buzz-worthy due to their ZSL ability.
However, they only care the performance on unseen datasets instead of unseen classes, and do not set any constraint on class splitting.
While it is unclear if they are inductive or transductive,  we position them as an additional type: dataset-level for an exhaustive comparison.

\subsubsection{Unknown Unseen Class Prior}
We first examine the non-ideal case that the GT unseen class prior $p(y^{u})$ is unknown.
we compare ours with the other methods under a naive uniform unseen prior and the advanced CPE estimation.
We report our results against existing ones. For methods needing unseen class prior, a uniform prior is used.
The results are reported in Table~\ref{tab:unknown}.
Evidently, across all priors, I-VAEGAN can demonstrate consistently better performance.
For example, under CPE prior, our method improves on non-uniform AWA1\&2 bigger than uniform CUB and SUN (2.6\% and 3.0\% v.s. 0.5\% and 1.8\%).
It is a strong evidence supporting our hypothesis: \textbf{directly using the pseudo unseen class semantics $\tilde{\mathbf{a}}^{u}$ predicted from real unseen class samples is better than estimating unseen class priors $p_g(y^{u})$ then sampling unseen class semantics.}

\subsubsection{Known Unseen Class Prior}
Next, we examine the more ideal situation, in which GT unseen class prior $p(y^{u})$ is available.
We focus on SOTAs under the same transductive ZSL setting for fair comparison.
We also compare the inductive ZSL methods for comprehensive reference and use T and I to distinguish TZSL or IZSL methods.

The TZSL results are reported in Table~\ref{tab:overall_zsl}.  The TGZSL results  are shown in Table~\ref{tab:overall_gzsl}.
As observed, we can conclude (a) For non-uniform datasets, our I-VAEGAN achieves great improvements. For example, in AWA1,  I-VAEGAN surpasses the second by 0.5$\%$ T1 for ZSL (ours 94.4$\%$ v.s. Bi-VAEGAN 93.9$\%$); and in AWA2 for TGZSL, I-VAEGAN improves 0.8$\%$ H (91.2$\%$ v.s. 90.4$\%$);
(b) For close-uniform datasets, I-VAEGAN obviously boosts the performance, e.g., at CUB, it raises $0.4\%$ for ZSL T1 and $0.6\%$ TGZSL H;
(c) Although dataset-based ZSL methods, i.e. CLIP and SHIP, use huge parameters and data,
their ZSL performance still generates a large gap compared to standard ZSL methods.

\subsection{Ablation Study}
\label{sec:ablation}
\subsection{Component Analysis}
\begin{table}[htbp]
    \centering
    \renewcommand{\multirowsetup}{\centering}
    \caption{Ablation study for TZSL.}
    
        \begin{tabular}{p{0.7cm}<{\centering}|p{0.7cm}<{\centering}|p{0.8cm}<{\centering}|p{0.75cm}<{\centering}p{0.75cm}<{\centering}p{0.75cm}<{\centering}p{0.75cm}<{\centering}}
        \hline
            \hline
            \multirow{1}{*}{PFA} & \multirow{1}{*}{VER} & \multirow{1}{*}{Prior} & AWA1$^\dagger$ & AWA2$^\dagger$ & CUB$^\ddagger$ & SUN$^\ddagger$ \\
            \hline
            $\times$  & $\times$  & Uni. & 66.3 & 58.8 & 77.0 & 75.8 \\
            $\times$  & $\times$ & CPE& 90.0 & 83.5 & 74.0 & 71.3 \\
            
            
            $\checkmark$ & $\times$ & CPE & 91.3 & 83.6 & 74.3 & 71.4 \\
            $\checkmark$ & $\checkmark$ & CPE & 92.6 & 86.5 & 74.5 & 73.1 \\
            
            $\checkmark$ & $\checkmark$ & GT & 94.4 & 95.9 & 77.2 & 76.0 \\
            \hline
            \hline
        \end{tabular}
    \label{tab:ablation}
\end{table}

The component analysis results for TZSL can be found at Table~\ref{tab:ablation}.
In the table, we report the ablated PFA and VER to verify the their effectiveness in ZSL.
Besides, it also displays the impact about prior type and the final classification space.
We can find that our VER and PFA lead a satisfactory performance gain.
We also provided the ablation study for TGZSL in \textbf{Appendix~\ref{appendix:addAblation}}.

\subsection{Hyper-parameters Analysis}
In our I-VAEGAN, the main hyper-parameters include the loss weight $\lambda_{u2}$ and the number of synthesized samples $n_{syn}$.
The results and analysis are provided \textbf{Appendix~\ref{appendix:addAblation}}.

\subsection{Further Analysis}
To further assess the effectiveness of PFA, we also evaluate the PFA under different priors.
The results are provided in \textbf{Appendix~\ref{appendix:additionalPFA}}.
We also draw the testing MAE loss comparison at testing dataset.
Besides, our VER can be easily integrated into other methods. E.g., we adopt it to TF-VAEGAN and FREE~\cite{chen2021free}.
Our VER can be easily integrated into existing methods, regardless inductive or transductive and significantly boosts regression accuracy for various ZSL methods.
These results are provided in \textbf{Appendix~\ref{appendix:additionalVER}}.

\subsection{Reduced APE}
\label{sec:APE}
\begin{figure}[htbp]
    \centering
    \includegraphics[width=\linewidth]{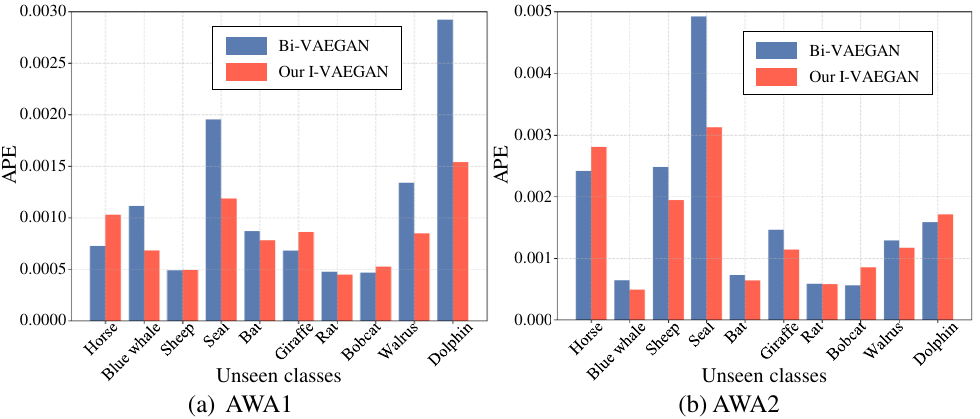}
    \caption{The APE comparison for our I-VAEGAN and Bi-VAEGAN on (a) AWA1 and (b) AWA2.
    Our method effectively reduces APE.}
    \label{fig:APE}
\end{figure}

To evaluate how well our method mitigates the proposed APE, we calculate per-class APE from Bi-VAEGAN and our I-VAEGAN, as shown in Fig.~\ref{fig:APE}.
Clearly, we can find that our method effectively reduces APE.

\subsection{Prior Estimation Comparison}
We further compare the CPE estimated prior to those from Bi-VAEGAN.
The results are provided in \textbf{Appendix~\ref{appendix:PriorEst}}.

\section{Conclusion}
In this paper, we revealed one potential issue for the unconditional unseen discriminator in standard f-VAEGAN, and proved it would accumulate prior bias, resulting in an inevitable class-specific generation bias. To tackle the problem, we propose a new architecture Improved VAEGAN (I-VAEGAN), based on two proposed simple but effective modules: Variational Embedding Regression (VER) and Pseudo-conditional Feature Adversarial (PFA).
VER effectively improves the attribute regression accuracy on unseen classes, while PFA reduces the prior bias by explicitly expressing class conditions.
Extensive experiments on four popular benchmarks show the effectiveness of our method.


\bibliography{IVAEGAN_arxiv}
\bibliographystyle{icml2025}

\newpage
\appendix
\onecolumn

\textbf{Appendix Organization}: We present additionally (\textbf{A}) the  proof for our Accumulated Prior Error Proposition, (\textbf{B}) the detailed training algorithm, (\textbf{C}) additional experiments  containing (\textbf{C1}) dataset information, (\textbf{C2}) implementation details, (\textbf{C3}) addition ablation study, (\textbf{C4}) PFA on various priors, (\textbf{C5}) VER on various methods, and (\textbf{C6}) prior estimation comparisons.

\section{Accumulated Prior Error Proposition}
\label{appendix:APE}

\begin{shaded}
    \begin{proposition}[APE]
    When G gets the expected global optimum of the minimax game, i.e., $ p_r(\mathbf{x}^u) = p_g(\mathbf{x}^u)$, for any unseen class $y^u_{i}$, 
        \begin{align}
        \label{eq:chain2}
            {\color{red} e(\mathbf{x}^u, y^u_i) } = |p_r(\mathbf{x}^u|y^u_{i}) - p_g(\mathbf{x}^u|y^u_{i})|,
        \end{align}
    where $e(\mathbf{x}^u, y^u_i)$ is the accumulated prior error:
        \begin{align}
        \label{eq:chainAE}
            & e(\mathbf{x}^u, y^u_i) = \frac{ | p_r(y^u_i|\mathbf{x}^u)p_g(y^u_i) - p_g(y^u_i|\mathbf{x}^u)p_r(y^u_i) | }{p_r(y^u_i)p_g(y^u_i)}  p_r(\mathbf{x}^u).
        \end{align}
    \end{proposition}
\end{shaded}

\begin{proof}
We can reformulate $|p_r(\mathbf{x}^u|y^u_{i}) - p_g(\mathbf{x}^u|y^u_{i})|$ by Bayes’ rule: 
\begin{align}
\label{eq:chain1_1}
& e(\mathbf{x}^u, y^u_i) = |p_r(\mathbf{x}^u|y^u_{i}) - p_g(\mathbf{x}^u|y^u_{i})| \\
\label{eq:chain1_2}
& = |\frac{p_r(\mathbf{x}^u) p_r(y_i|x^u) }{p_r(y^u_i)} - \frac{p_g(\mathbf{x}^{u}) p_g(y_i|\mathbf{x}^u) }{p_g(y^u_i)}|
\end{align}
Here $p_r(y^u_{i})$ is the ground truth prior probability for unseen class $i$, and $p_g(y^u_{i})$ is the supposed probability for the $i$-th unseen class from our selected prior estimation or pre-defined prior distribution.
Combining $p_r(\mathbf{x}) = p_g(\mathbf{x})$ at the global optimum, we can rewrite Eq.\ref{eq:chain1_1} as:
\begin{align}
    \label{eq:chain2_1}
    & |\frac{p_r(\mathbf{x}^u) p_r(y^u_i|x^u) }{p_r(y^u_i)} - \frac{p_g(\mathbf{x}^{u}) p_g(y^u_i|\mathbf{x}^u) }{p_g(y^u_i)}| \\
    \label{eq:chain2_2}
     & = |\frac{p_r(y^u_i|x^u) }{p_r(y^u_i)} - \frac{p_g(y^u_i|\mathbf{x}^u) }{p_g(y^u_i)}| p_r(\mathbf{x}^u) \\
    \label{eq:chain2_3}
    & = |\frac{p_r(y^u_i|\mathbf{x}^u)p_g(y^u_i) - p_g(y^u_i|\mathbf{x}^u)p_r(y^u_i) }{p_r(y^u_i)p_g(y^u_i)} | p_r(\mathbf{x}^u) \\
    \label{eq:chain2_4}
    & = \frac{ | p_r(y^u_i|\mathbf{x}^u)p_g(y^u_i) - p_g(y^u_i|\mathbf{x}^u)p_r(y^u_i) | }{p_r(y^u_i)p_g(y^u_i)}  p_r(\mathbf{x}^u).
\end{align}

\end{proof}

\begin{remark}
    $p_r(y^u_{i}|\mathbf{x}^{u})$ is the real unseen-class-posterior probability, while $p_g(y^u_{i}|\mathbf{x}^{u})$ is the generated unseen-class-posterior probability.
    The $e(\mathbf{x}^{u}, y^u_i)$ becomes $0$ if $| p_r(y^u_i|\mathbf{x}^u)p_g(y^u_i) - p_g(y^u_i|\mathbf{x}^u)p_r(y^u_i) | = 0$, i.e.,
    \begin{align}
    \frac{p_r(y^u_i|\mathbf{x}^u)}{p_g(y^u_i|\mathbf{x}^u) } = \frac{p_r(y^u_i)}{p_g(y^u_i)}.
    \end{align}
    In other words, only when the classification probabilities $p_r(y^u_i|\mathbf{x}^u)$ (trained by real data) and $p_g(y^u_i|\mathbf{x}^u)$ (trained by fake generated data) are perfectly in proportion to the real class probability $p_r(y^u_i)$ and estimated class probability $p_g(y^u_i)$.
    However, existing generation methods still far away from this point as the quantitative comparison of APE we provided in experiments.
    Therefore, the issue of unconditional $D_{u}$ cannot be avoided, showing that $D_{u}$ might be the key factor.
    To this end, a prior reaction chain is emerged, as shown in Fig.~\ref{fig:chain}:
    \begin{align}
    \label{eq:chain}
        p_g(y^{u}) \to D_{u} \to G \to f_{zsl},
    \end{align}
\end{remark}

\section{Algorithm}
\label{appendix:alg}

Our complete training algorithm is outlined in Alg.~\ref{alg:ivaegan}, which contains three stages.
The first stage involves pre-training a Variational Auto-Encoding (VAE), with its encoder and decoder denoted as $E_{pre}$ and $F_{pre}$, respectively.
The second stage focuses on training our visual-semantic regressor $R$ and the semantic discriminator $D_{r}$.
The third stage trains another encoder $E$, generator $G$, seen discriminator $D_{s}$, unconditional unseen discriminator $D_{u1}$,  and conditional unseen discriminator $D_{u2}$.
The first stage is trained independently, while the last two stages are trained alternatively.
Besides, we provide all the loss functions and network definitions in this section.

\subsection{Stage-1: Pre-training}
Specifically, let $E_{pre}$ denote the visual encoder and $F_{pre}$   denote the decoder.
We pretrain these components via a standard VAE loss function:
\begin{align}
\label{appeq:ver}
\mathcal{L}^{pre}_{VAE} & = KL\left( \mathcal{N}(\mathbf{\mu}_{pre}, \mathbf{\sigma}_{pre}) \| \mathcal{N}(\mathbf{0},\mathbf{1}) \right) \nonumber \\
&+ \mathbb{E}_{ \tilde{\mathbf{z}} \sim \mathcal{N}(\mathbf{\mu}_{pre}, \mathbf{\sigma}_{pre}) } [\| F_{pre}(\tilde{\mathbf{z}}) -  \mathbf{x}^{s/u}\|^2_2 ].
\end{align}
Next, we fix $E_{pre}$, and use its variational embedding to construct a low-dimensional latent representation space for all classes.
In the subsequent regression step, the variational embedding $\mathbf{\mu}_{pre}, \mathbf{\sigma}_{pre}$ are concatenated with the original visual features.
The VER then uses the concatenated vector $[ \mathbf{x}, \mathbf{\mu}_{pre}, \mathbf{\sigma}_{pre} ]$ to replace the original features, thus integrating with other two types of regressions.

\subsection{Stage-2: Regressor Training}
Using our pre-trained $E_{pre}$, we employ its variational embedding with a plain MSE loss and adversarial regression to train our regressor $R$.
We denote $\tilde{\mathbf{a}}^s = R([\mathbf{x}^s, \mathbf{\mu}^s_{pre}, \mathbf{\sigma}^s_{pre} ]) $ and $\tilde{\mathbf{a}}^u = R([\mathbf{x}^u, \mathbf{\mu}^u_{pre}, \mathbf{\sigma}^u_{pre} ]) $.
The MSE loss with VER is computed based on these predictions:
\begin{equation}
\label{appeq:mse}
\mathcal{L}_{mse} =\|  \mathbf{a}^s - \tilde{\mathbf{a}}^s \|^2 .
\end{equation}

The adversarial regression loss with VER is written as follows:
\begin{align}
    \label{appeq:gan_r}
    \mathcal{L}^{r}_{GAN} & = \mathbb{E}[D_r(\mathbf{a}^u)] - \mathbb{E}[D_r(\tilde{\mathbf{a}}^u)] - \lambda_{gp} \mathbb{E}[(\|\nabla_{\hat{\mathbf{a}}^u} D_r(\hat{\mathbf{a}}^u)\|_2 - 1)^2] \nonumber \\ 
    & + \mathbb{E}[D_r(\mathbf{a}^s)] - \mathbb{E}[D_r(\tilde{\mathbf{a}}^s)] - \lambda_{gp} \mathbb{E}[(\|\nabla_{\hat{\mathbf{a}}^s} D_r(\hat{\mathbf{a}}^s)\|_2 - 1)^2].
\end{align}

\subsection{Stage-3: Generator Training}
Following the approach of f-VAEGAN, our I-VAEGAN also combines a VAE and a Generative Adversarial Network (GAN).
Note that the VAE used in I-VAEGAN is different from the one pre-trained in stage-1.
The VAE loss for our model is given by Eq.~\ref{eq:fvaegan-vae}:

\begin{align}
\label{appeq:fvaegan-vae}
\mathcal{L}^{s}_{VAE} & = KL\left( \mathcal{N}(\mathbf{\mu}^s,\mathbf{\sigma}^s) \| \mathcal{N}(\mathbf{0},\mathbf{1}) \right) \nonumber \\
&+ \mathbb{E}_{ \tilde{\mathbf{z}}^{s} \sim \mathcal{N}(\mathbf{\mu}^s,\mathbf{\sigma}^s) } [\| G(\tilde{\mathbf{z}}^s, \mathbf{a}^{s}) -  \mathbf{x}^{s}\|^2_2 ].
\end{align}

It uses a conditional GAN loss for seen classes, given by Eq.~\ref{appeq:fvaegan-gan-s}, and an unconditional GAN loss for unseen classes, given by Eq.~\ref{appeq:fvaegan-gan-u}.

\begin{align}
\label{appeq:fvaegan-gan-s}
    \mathcal{L}^{s}_{GAN} &= \mathbb{E}[D_s(\mathbf{x}^s, \mathbf{a}^{s})] - \mathbb{E}[D_s(\tilde{\mathbf{x}}^s, \mathbf{a}^{s})] \nonumber \\
    & - \lambda_{gp} \mathbb{E}[(\|\nabla_{\hat{\mathbf{x}}^s} D_s(\hat{x}^s, \mathbf{a}^{s})\|_2 - 1)^2].
\end{align}

\begin{align}
\label{appeq:fvaegan-gan-u}
    \mathcal{L}^{u1}_{GAN} &= \mathbb{E}[D_u(\mathbf{x}^u)] - \mathbb{E}[D_u(\tilde{\mathbf{x}}^u)] \nonumber \\
    & - \lambda_{gp} \mathbb{E}[(\|\nabla_{\hat{\mathbf{x}}^u} D_u(\hat{\mathbf{x}}^u)\|_2 - 1)^2].
\end{align}

We propose the pseudo-conditional feature adversarial (PFA) loss:
\begin{align}
\label{appeq:pfa}
    \mathcal{L}^{u2}_{GAN} &= \mathbb{E}[D_{u2}(\mathbf{x}^u, \tilde{\mathbf{a}}^u)] - \mathbb{E}[D_{u2}(\tilde{\mathbf{x}}^u, \tilde{\mathbf{a}}^u)] \nonumber \\
    & - \lambda_{gp} \mathbb{E}[(\|\nabla_{\hat{\mathbf{x}}^u} D_{u2}(\hat{\mathbf{x}}^u, \tilde{\mathbf{a}}^u)\|_2 - 1)^2].
\end{align}

\subsection{Overall Objective}
Formally, the overall objective function of I-VAEGAN is formulated as:
\begin{align}
    \label{appeq:overall}
    & \text{For stage-1: } \min_{E_{pre}, F_{pre}} \mathcal{L}^{pre}_{VAE} \nonumber \\
    & \text{For stage-2: } \min_{R} \max_{D_{r}} \mathcal{L}_{mse}+\lambda_{r} \mathcal{L}^{r}_{GAN} \\
    & \text{For stage-3: } \min_{E,G} \max_{D_s,D_u,D_{u2}} (L_{VAE}^s +L_{GAN}^s  + L_{GAN}^u). \nonumber
\end{align}
Here $ L_{GAN}^u = \lambda_{u1} L_{GAN}^{u1}  +\lambda_{u2} L_{GAN}^{u2}$, and $ \lambda_{u1} $, $ \lambda_{u2} $, $ \lambda_{r} $ are hyper-parameters that control the influence of $ L_{GAN}^{u1}$ , $ L_{GAN}^{u2}$ and $ L_{GAN}^{r}$  on the generator.

\begin{algorithm}[htbp]
\caption{I-VAEGAN Training}
\label{alg:ivaegan}
\textbf{Input}: Training set $D^{tr} = \{\langle \mathcal{X}^{s}, \mathcal{Y}^s\rangle, \mathcal{X}^{u}, \{\mathcal{A}^s, \mathcal{A}^u\} \}$, $\lambda_{gp}$, $\lambda_{r}$, $\lambda_{u1}$, $\lambda_{u2}$, $n_{syn}$, the epoch numbers for pre-training $n_{pre}$, regressor training $n_{r}$ and generator training $n_{g}$. \\

\textbf{Output}: Semantic regressor $R$ and feature generator $G$.

\begin{algorithmic}[1] 
\STATE \textbf{Stage-1: VER Pre-training.}
\STATE Define an encoder $E_{pre}$ and a decoder $F_{pre}$ of VER;

\FOR{$i = 1$ to $n_{pre}$}
    \STATE Sample visual features $\mathbf{x^{s}}$ and $\mathbf{x^{u}}$ from $D^{tr}$;
    \STATE Project features to variational embedding $[\mathbf{\mu}_{pre}, \mathbf{\sigma}_{pre}] \leftarrow E_{pre}(\mathbf{x})$ and $\tilde{\mathbf{z}} \sim \mathcal{N}(\mathbf{\mu}_{pre}, \mathbf{\sigma}_{pre})$;
    \STATE Reconstruct visual features by $\tilde{\mathbf{x}} \leftarrow F_{pre}(\tilde{\textbf{z}})$;
    \STATE Update $E_{pre}$ and $F_{pre}$ by minimizing $\mathcal{L}^{pre}_{VAE}$ (Eq.~\ref{appeq:ver}); 
\ENDFOR
\STATE Freeze $E_{pre}$;

\STATE \textbf{Stage-2: Regressor Training}
\STATE Define semantic regressor $R$ and unconditional semantic discriminator $D_{r}$;
\FOR{$i = 1$ to $n_r$}
    \STATE Sample a batch of visual features $\mathbf{x}^{s}$ and their corresponding semantic labels $\mathbf{a}^{s}$;
    \STATE Sample a batch of semantic labels $\mathbf{a}^{u}$ according to predefined prior estimation;
    \STATE Project $\mathbf{x^{s}}$ and $\mathbf{x^{u}}$ to VER embedding space $[\mathbf{\mu}_{pre}, \mathbf{\sigma}_{pre}] \leftarrow E_{pre}(\mathbf{x})$;
    \STATE Predict semantic label $\mathbf{\tilde{a}} = R([ \mathbf{x},\mathbf{\mu}_{pre}, \mathbf{\sigma}_{pre}]$;
    \STATE Update $D_{r}$ by maximizing $\mathcal{L}_{mse}$ (Eq.~\ref{appeq:gan_r}) ;
    \STATE Update $R$ by minimizing  $\mathcal{L}_{mse}$ (Eq.~\ref{appeq:gan_r}) and $\mathcal{L}^{r}_{GAN}$ (Eq.~\ref{appeq:mse});
\ENDFOR
\STATE Freeze $R$;

\STATE \textbf{Stage-3: Generator Training}
\STATE Define feature encoder $E$, generator $G$, conditional seen discriminator $D_{s}$, unconditional unseen discriminator $D_{u}$ and conditional unseen discriminator $D_{u2}$;
\FOR{$i = 1$ to $n_g$}
    \STATE Sample a batch of visual features $\mathbf{x}^{s}$ and their corresponding semantic labels $\mathbf{a}^{s}$;
    \STATE Sample a batch of unlabeled visual features $\mathbf{x}^{u}$;
    \STATE Project $\mathbf{x}^{s}$ and $\mathbf{a}^{s}$ to latent space $[\mathbf{\mu}^{s}, \mathbf{\sigma}^{s}] \leftarrow E(\mathbf{x}^{s},\mathbf{a}^{s})$ and $\tilde{\mathbf{z}}^{s} \sim \mathcal{N}(\mathbf{\mu}_{pre}, \mathbf{\sigma}_{pre})$;
    \STATE Generate fake features of seen class $\tilde{\mathbf{x}}^{s} \leftarrow G( \tilde{\mathbf{z}}^{s}, \mathbf{a}^{s})$;
    
    \STATE Generate fake features of unseen class $\tilde{\mathbf{x}}^{u} \leftarrow G( \mathbf{z}, \tilde{\mathbf{a}}^{u})$;
    \STATE Update $D_{s}$ by maximizing $\mathcal{L}^{s}_{GAN}$ (Eq.~\ref{appeq:fvaegan-gan-s}); 
    \STATE Update $D_{u}$ by maximizing $\mathcal{L}^{u1}_{GAN}$ (Eq.~\ref{appeq:fvaegan-gan-u}); 
    \STATE Update $D_{u2}$ by maximizing $\mathcal{L}^{u2}_{GAN}$ (Eq.~\ref{appeq:pfa}); 

    \STATE Update $E$ and $G$ by minimizing 
$\mathcal{L}^{s}_{VAE}$ (Eq.~\ref{appeq:fvaegan-vae}), $\mathcal{L}^{s}_{GAN}$ (Eq.~\ref{appeq:fvaegan-gan-s}), $\mathcal{L}^{u1}_{GAN}$ (Eq.~\ref{appeq:fvaegan-gan-u}) and $\mathcal{L}^{u2}_{GAN}$ (Eq.~\ref{appeq:pfa});
\ENDFOR
\STATE Freeze $G$.

\end{algorithmic}
\end{algorithm}

\begin{figure*}[htbp]
    \centering
    \includegraphics[width=\linewidth]{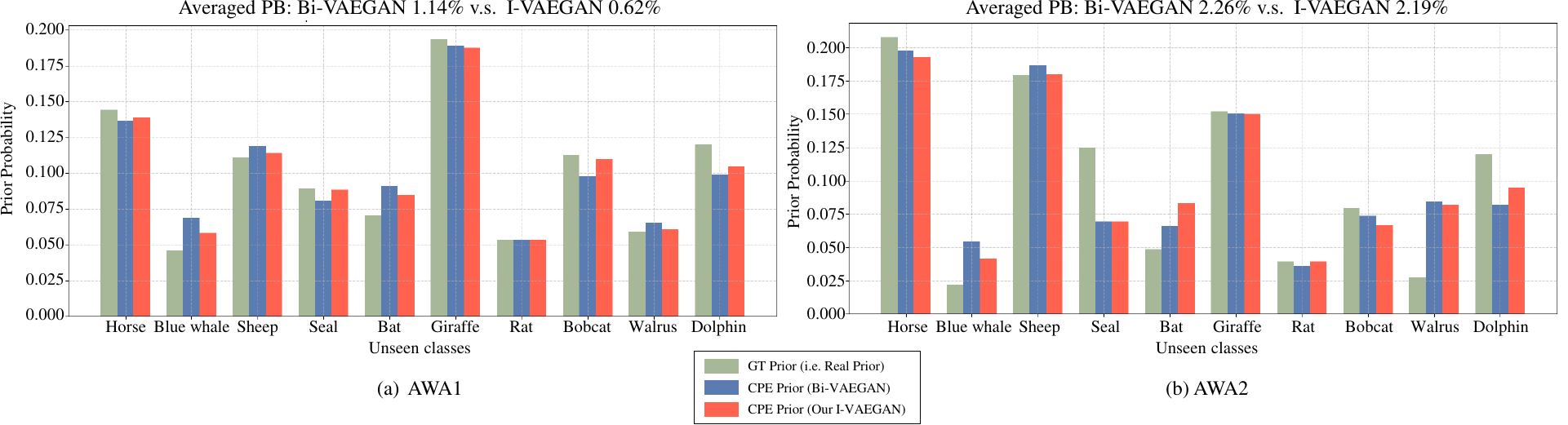}
    \caption{The comparison for prior estimation in non-uniform datasets: AWA1 and AWA2.}
    \label{fig:PB_bar1}
\end{figure*}

\begin{figure*}[htbp]
    \centering
    \includegraphics[width=\linewidth]{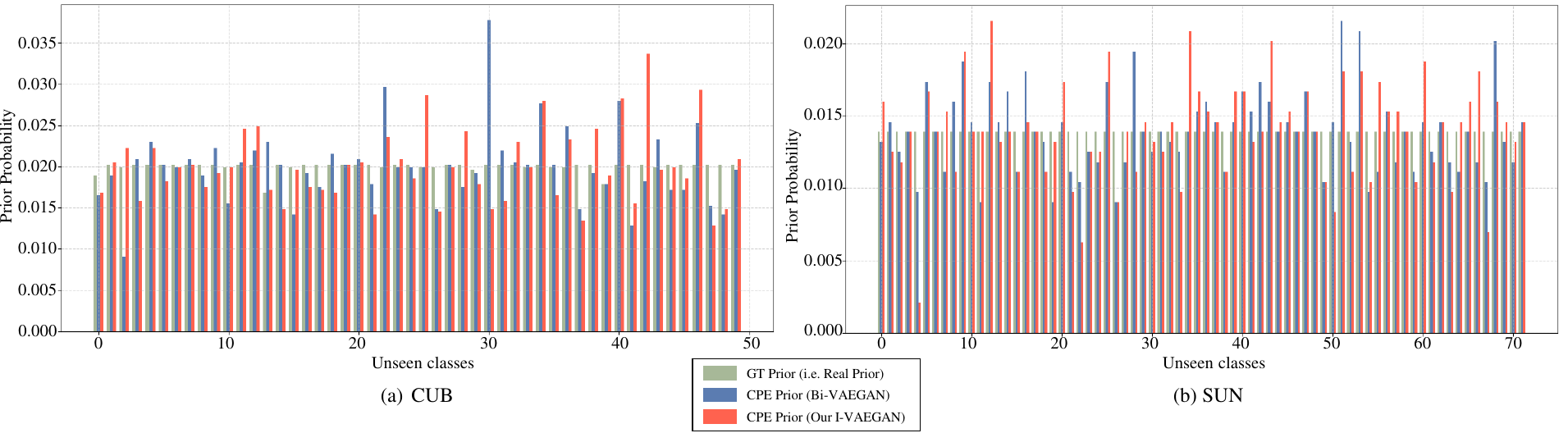}
    \caption{The comparison for prior estimation in close-uniform datasets: CUB and SUN.}
    \label{fig:PB_bar2}
\end{figure*}

\begin{figure*}[htbp]
    \centering
    \includegraphics[width=\linewidth]{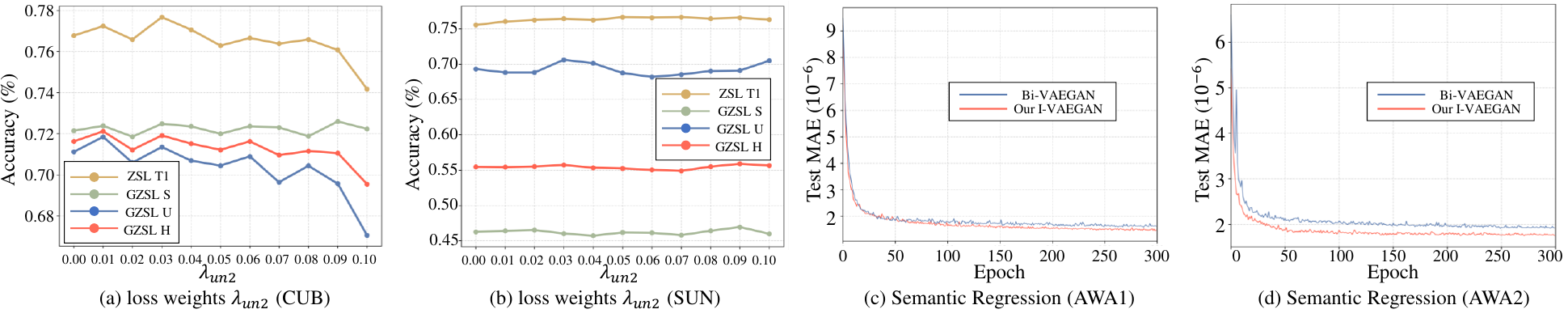}
    \caption{Accuracy curve on (a) AWA1 and (b) AWA2 and error curve on (c) AWA1 and (d) AWA2.}
    \label{fig:curve}
\end{figure*}

\begin{figure*}[htbp]
    \centering
    \includegraphics[width=\linewidth]{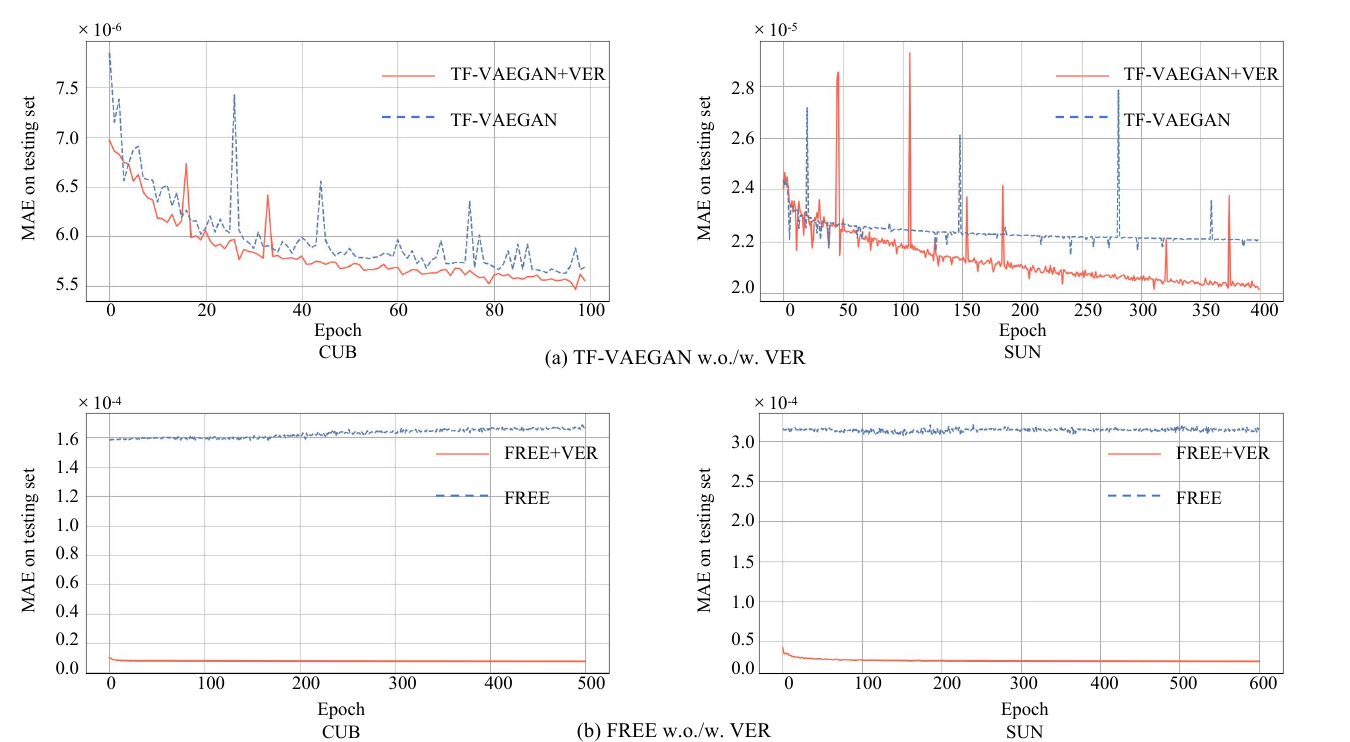}
    \caption{Our VER improves semantic regression across various existing methods.}
    \label{fig:ver_compare}
\end{figure*}

\begin{figure*}[htbp]
    \centering
    \includegraphics[width=\linewidth]{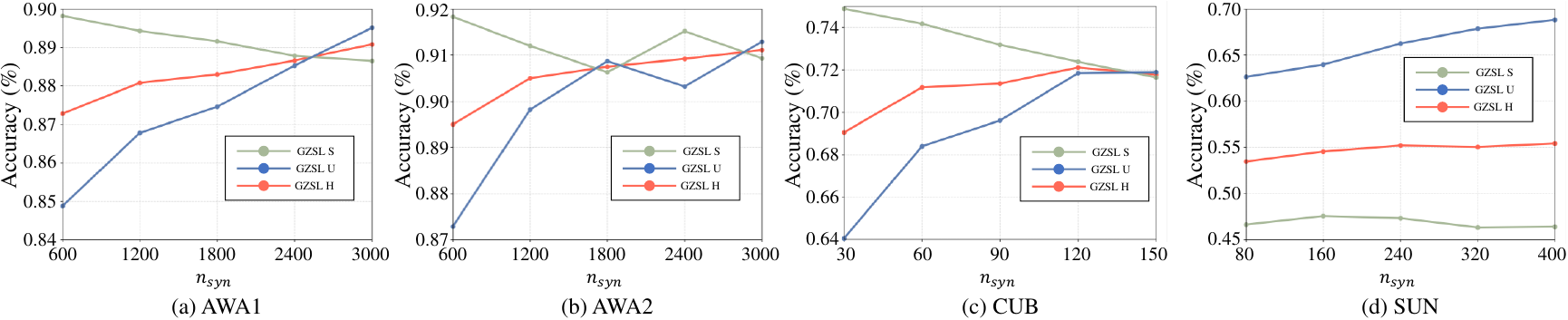}
    \caption{Varying the number of synthesized samples of unseen classes.}
    \label{fig:num_syn}
\end{figure*}

\section{Additional Experiments}
\label{appendix:addExp}

\subsection{Additional Dataset Information}
\label{appendix:addDataset}

To demonstrate TZSL's ability of our I-VAEGAN, we conduct experiments using four popular benchmark datasets, including two \textbf{non-uniform ($\dagger$)} animal datasets: AWA1~\cite{lampert2013attribute} and AWA2~\cite{xian2018zero}, and two \textbf{close-uniform ($\ddagger$)} datasets: CUB~\cite{wah2011caltech} for birds and SUN~\cite{patterson2012sun} for scenes.
Visual features are extracted using ResNet101~\cite{he2016deep} pre-trained on ImageNet~\cite{deng2009imagenet}.
We partition these data into training and testing sets following~\cite{narayan2020latent, xian2018zero}, which are widely used in current approaches to ensure fair comparison.
The partitioning guarantees that unseen classes are rigorously excluded from ImageNet.

\subsection{Implementation Details}
\label{appendix:imp}

For a fair comparison, we follow Bi-VAEGAN~\cite{wang2023bi} using the AdamW optimizer~\cite{loshchilov2017decoupled} with a learning rate of 0.001 and parameters $\beta_1 = 0.5$, $\beta_2 = 0.999$.
We set the mini-batch size to 64 for all datasets.
The training epochs are set to 300 for AWA1, AWA2, 600 for CUB and 400 for SUN.

During the inference stage, the number of synthesized features per class is set to 3,000 for AWA1 and AWA2, 120 for CUB, and 400 for SUN, respectively.
Our experiments are implemented using PyTorch 3.8~\cite{paszke2019pytorch}on an NVIDIA GeForce RTX 3090 GPU.

The encoder $E_{pre}$ and decoder $D_{pre}$ in the $VER$, as well as feature encoder $E$, generator $G$ and regressor $R$ in I-VAEGAN are all implemented as two-layer Multi-Layer Perceptrons (MLPs).
Each hidden layer has 4,096 dimensions, and the activation function used is LeakyReLU.
The conditional seen discriminator $D_{s}$, unconditional unseen discriminators $D_{u}$ and $D_{u2}$, and the semantic discriminator $D_{r}$ are also two-layer MLPs, with the final output layer producing a scalar value.
The final classifier $f_{zsl}$ is a single fully-connected layer, with its output dimension corresponding to the number of unseen classes for TZSL or the total number of both seen and unseen classes for transductive GZSL (TGZSL).

\begin{table*}[htbp]
    \centering
    \renewcommand{\multirowsetup}{\centering}
\caption{The ablation study for transductive GZSL evaluates our proposed VER, PFA and different prior. $\dagger$ and $\ddagger$ indicate datasets with non-uniform or close-uniform distributions of unseen classes, respectively.
    In GZSL, U, S and H represent the top-1 accuracy (\%) of unseen classes, seen classes, and their harmonic mean, respectively.}
    
        \begin{tabular}{c|c|c|ccc|ccc|ccc|ccc}
        \hline
            \hline
            \multirow{2}{*}{\textbf{PFA}} & \multirow{2}{*}{\textbf{VER}} & \multirow{2}{*}{\textbf{Prior}} & \multicolumn{3}{c|}{AWA1$^\dagger$} & \multicolumn{3}{c|}{AWA2$^\dagger$} & \multicolumn{3}{c|}{CUB$^\ddagger$} & \multicolumn{3}{c}{SUN$^\ddagger$} \\
            \cline{4-15}
            & & & U & S & H & U & S & H & U & S & H & U & S & H \\
            \hline
            $\times$  & $\times$  & Uni.& 58.4 & 74.3 & 65.4 & 51.1 & 80.5 & 62.5 & 74.1 & 66.2 & 69.9 & 70.8 & 45.4& 54.4 \\
            $\times$  & $\times$  & CPE & 86.9 & 79.7 & 83.1 & 79.2 & 82.4 & 80.8 & 71.2 & 66.2 & 68.6 & 65.1 & 45.7 & 53.7 \\
            
            $\checkmark$ & $\times$ & CPE & 86.1 & 87.5 &86.8 & 77.7 & 84.9 & 81.1 & 67.8 & 71.8 & 69.8 & 66.3 & 45.7 & 54.1 \\
            $\checkmark$ & $\checkmark$ & CPE & 86.7 &88.8 & 87.8 & 80.7 & 86.5 & 83.5 & 69.8 & 71.2 & 70.5 & 66.6 & 45.8 & 54.3 \\
            
            $\checkmark$ & $\checkmark$ & GT  & 89.5 & 88.6 & 89.1 & 91.2 & 91.3 & 91.2 & 71.8&72.3&72.1 & 68.8 & 46.4 & 55.4 \\
            \hline
            \hline
        \end{tabular}
    
    \label{tab:ablation_GZSL}
\end{table*}

\begin{table*}[thp]
    \centering
    \footnotesize
    \renewcommand{\multirowsetup}{\centering}
    \caption{Our PFA improves performance across varied priors in transductive GZSL generally (the classification is performed solely in the visual space).
    The symbols $\dagger$ and $\ddagger$ denote datasets with non-uniform or close-uniform distribution of unseen classes, respectively.
    In GZSL, U, S and H represent the top-1 accuracy (\%) of unseen classes, seen classes, and their harmonic mean, respectively.}

        \begin{tabular}{c|c|ccc|ccc|ccc|ccc}
            \hline
            \hline
            \multirow{2}{*}{\textbf{Prior}} & \multirow{2}{*}{\textbf{PFA}} & \multicolumn{3}{c}{AWA1$^\dagger$} & \multicolumn{3}{|c}{AWA2$^\dagger$} & \multicolumn{3}{|c|}{CUB$^\ddagger$} & \multicolumn{3}{c}{SUN$^\ddagger$} \\
            \cline{3-14}
            & & U & S & H & U & S & H & U & S & H & U & S & H \\
            \hline
            \multirow{2}{*}{Uni.}  & $\times$ & \textbf{70.4} & 33.0 & 44.9 & 60.7 & \textbf{47.0} & 53.0 & \textbf{60.8} & 36.2 & 45.4 & \textbf{58.5} & 31.4 & 40.9 \\
             & $\checkmark$ & 60.2 & \textbf{39.8} & \textbf{48.0} & \textbf{65.3} & 45.2 & \textbf{53.4} & 53.5 & \textbf{47.4} & \textbf{50.3} & 53.4 & \textbf{38.9} & \textbf{45.0} \\
            \hline
            \multirow{2}{*}{CPE}  & $\times$ & \textbf{70.4} & 32.3 & 44.3 & 62.3 & \textbf{47.8} & \textbf{54.0} & \textbf{60.1} & 36.3 & 45.3 & \textbf{59.3} & 31.3 & 41.0 \\
             &  $\checkmark$  & 61.5 & \textbf{39.9} & \textbf{48.4} & \textbf{65.2} & 45.2 & 53.4 & 51.8 & \textbf{49.7} & \textbf{50.7} & 50.4 & \textbf{40.8} & \textbf{45.1} \\
            \hline
            \multirow{2}{*}{GT}  & $\times$ & \textbf{69.9} & 31.8 & 43.8 & \textbf{65.7} & \textbf{45.5} & \textbf{53.7} & \textbf{59.6} & 36.2 & 45.1 & \textbf{58.8} & 32.4 & 41.7 \\
             &  $\checkmark$ & 63.7 & \textbf{38.4} & \textbf{47.9} & 65.1 & 44.8 & 53.1 & 52.2 & \textbf{49.6} & \textbf{50.8} & 53.0 & \textbf{39.3} & \textbf{45.2} \\
            \hline
            \hline
        \end{tabular}
    
    \label{tab:effect_PFA_GZSL}
\end{table*}

\subsection{Additional Ablation Study}
\label{appendix:addAblation}

\subsubsection{Component Analysis for TGZSL}
We provide the component analysis on transductive GZSL (TGZSL) in Table~\ref{tab:ablation_GZSL}.
The results demonstrate that  PFA and VER consistently improve the performance for both non-uniform and close-uniform datasets.
For example, on the non-uniform AWA1 dataset, our PFA significantly enhances H from 83.1\% to 86.1\% , while VER further increases H by 1.0\%.
On the close-uniform CUB dataset, our PFA boosts H from 68.6\% to 69.8\%, and VER raises H to 70.5\%.

\subsubsection{Hyper-parameters Analysis}
In our I-VAEGAN, the main hyper-parameters include the loss weight $\lambda_{u2}$ and the number of synthesized samples $n_{syn}$.
We primarily tune $\lambda_{u2}$ (as defined in Eq.~\ref{eq:overall}).
The sensitivity analysis of $\lambda_{u2} $ is illustrated in Fig.~\ref{fig:curve} (a) and (b).
The sensitivity varies across different datasets.
For CUB, the performance changes significantly with different values of $\lambda_{u2}$, with smaller values leading to better performance.
In contrast, for SUN, performance changes only slightly as $\lambda_{u2}$ increases.
The best performance is achieved when $\lambda_{u2}$ is set to 0.09, maximizing both H and T1.

The impact of $n_{syn} $ is illustrated in Fig.~\ref{fig:num_syn}.
The accuracy for unseen class varies with the number of synthesized samples, with the performance reaching its peak when $n_{syn} = 3000$.
This result demonstrates that the features synthesized by our method effectively mitigate the issue of missing data for unseen classes.

\subsection{PFA for Various Priors}
\label{appendix:additionalPFA}

We further analyze our PFA across different priors, as shown in Tab.~\ref{tab:effect_PFA_GZSL}.
Our results indicate that PFA consistently achieves better performance in most cases.
For instance, with a uniform prior, our PFA improves H from 44.9\% to 48.0\% on AWA1.
When using the CPE prior, our PFA boosts H from 45.3\% to 50.7\% on CUB.
Additionally, with the Ground Truth (GT) prior, our PFA has a 3.5\% improvement in H on SUN.

\subsection{VER on Various Methods}
\label{appendix:additionalVER}

We plot the results of semantic regression on AWA1 and AWA2, as shown in Fig.~\ref{fig:curve} (c) and (d).
Besides, our VER can be easily integrated into existing methods, regardless inductive or transductive.
We add our VER into TF-VAEGAN~\cite{narayan2020latent} and FREE~\cite{chen2021free} to verify this.
The results are presented in Fig.~\ref{fig:ver_compare}, where a significant reduction  is observed for semantic Mean Absolute Error (MAE) on test set.

\subsection{Prior Estimation Comparison}
\label{appendix:PriorEst}

We compare Bi-VAEGAN~\cite{wang2023bi} with our I-VAEGAN in terms of CPE prior estimation.
The results on non-uniform datasets AWA1 and AWA2 are shown in in 
 Fig.~\ref{fig:PB_bar1}, while the results for close-uniform datasets CUB and SUN are displayed in 
 Fig.~\ref{fig:PB_bar2}.
Our I-VAEGAN demonstrates a lower class-average Prior Bias (PB) on non-uniform datasets.
For instance, the averaged PB of Bi-VAEGAN is 1.14\%, whereas ours is 0.62\%.

\end{document}